\documentclass[11pt,a4paper]{article}

\usepackage[utf8]{inputenc}
\usepackage[T1]{fontenc}
\usepackage{amsmath,amssymb,amsfonts,amsthm}
\usepackage{pifont}
\usepackage{graphicx}
\usepackage{booktabs}
\usepackage{multirow}
\usepackage{hyperref}
\usepackage{cleveref}
\newtheorem{definition}{Definition}
\newtheorem{property}{Property}
\crefname{property}{Property}{Properties}
\Crefname{property}{Property}{Properties}
\usepackage{algorithm}
\usepackage{algorithmic}
\usepackage{xcolor}
\usepackage{enumitem}
\usepackage{natbib}
\usepackage[margin=1in]{geometry}
\usepackage{subcaption}
\usepackage{tabularx}
\usepackage{tikz}
\usetikzlibrary{trees,arrows.meta,positioning}

\newif\ifcodereleased
\codereleasedfalse

\newcommand{\sct}{\textsc{SCT}}
\newcommand{\residual}{\mathcal{R}}
\newcommand{\compress}{\mathcal{C}}

\title{Semantic Compression Trees: Multi-Resolution Knowledge Retrieval\\via Hierarchical Semantic Residuals}

\author{
  Junaid Farooq\\[4pt]
  Sprouts.ai\\
  \href{mailto:junaid.farooq@sprouts.ai}{\texttt{junaid.farooq@sprouts.ai}}\\[4pt]
  National Institute of Technology Srinagar\\
  \href{mailto:junaid_phd017@nitsri.ac.in}{\texttt{junaid\_phd017@nitsri.ac.in}}\\[4pt]
  \href{https://junaidfarooq.net}{junaidfarooq.net}
}

\begin{document}
\maketitle

\begin{abstract}
Retrieval-augmented generation relies mostly on flat, fixed-granularity indexes:
documents are cut into uniform chunks and retrieved by similarity, discarding the
hierarchical structure of the source. We introduce \textbf{Semantic Compression
Trees (\sct{})}, a hierarchical index in which each node stores only its
\emph{semantic residual} --- the information it adds beyond its parent --- and
retrieval proceeds by progressive descent from the root, so that per-query cost is
governed by tree depth rather than collection size.

We evaluate on QASPER~\citep{dasigi2021qasper} (50 papers, 173 questions) under
two protocols differing only in whether the benchmark supplies the relevant
document, with bootstrap confidence intervals and paired significance tests
throughout. The results are mixed and we report them as such. When the document is
given, \sct{} with a zero-LLM extractive compressor matches dense retrieval on
answer quality (0.274 vs.\ 0.277 F1, $p = 0.37$) using \textbf{30\% fewer context
tokens} and \textbf{no LLM calls} to build the index, and residual storage beats
storing full summaries at each node (0.274 vs.\ 0.205, $p < 0.001$). Increasing the
collection fifty-fold multiplies flat retrieval's per-query scoring work by
$48.9\times$ and \sct{}'s by $6.4\times$.

Progressive descent itself is not supported. Retrieving the same residuals
\emph{without} the tree performs identically when the document is given
($p = 0.27$), and descent is substantially worse when the system must select the
document (0.122 vs.\ 0.165, $p < 0.001$). Routing accuracy localises the cause:
descent selects the correct paper 20.2\% of the time against 39.3\% for flat
retrieval, because that choice is made from the root residual, the most compressed
node in the tree. We conclude that the residual representation is worth keeping
and top-down routing is not.
\end{abstract}

\section{Introduction}
\label{sec:intro}

Retrieval-Augmented Generation (RAG) has become the dominant paradigm for grounding large language models (LLMs) in external knowledge~\citep{lewis2020rag,guu2020realm}. The standard pipeline---chunk documents into fixed-size segments, embed them into a dense vector space~\citep{karpukhin2020dpr}, and retrieve the top-$k$ most similar chunks at query time---is conceptually simple and widely deployed~\citep{gao2024ragsurvey}. However, this flat retrieval approach has three fundamental limitations that we address in this work.

\textbf{Similarity $\neq$ relevance.} Dense retrieval models measure semantic proximity in embedding space~\citep{karpukhin2020dpr,izacard2022contriever,khattab2020colbert}, but proximity does not imply informational relevance. A passage about ``temperature changes in the Arctic'' is semantically \emph{closer} to a query about ``climate in Antarctica'' than a passage about ``policy responses to Arctic warming,'' yet the latter may be more \emph{relevant} for answering ``What policies address Arctic warming?'' This distinction between similarity and relevance is well-documented in information retrieval~\citep{thakur2021beir,zhao2022densesurvey}.

\textbf{Fixed granularity is suboptimal.} Some queries require only a high-level summary (``What is this paper about?'') while others demand specific details (``What F1 score did Model X achieve on dataset Y?''). Fixed-size chunking treats all queries identically. Recent work on adaptive chunking~\citep{zhong2024mixgranularity,gunther2024latechunking} and hierarchical chunking~\citep{hichunk2025} has begun to address this, but these approaches still operate within a flat retrieval paradigm.

\textbf{Document structure is discarded.} Documents have inherent hierarchical organization---chapters, sections, subsections, paragraphs---that encodes semantic relationships and scoping. Traditional RAG pipelines destroy this structure during chunking~\citep{chunking2025eval}. The Hierarchical Attention Network~\citep{yang2016han} demonstrated the value of preserving document hierarchy for classification; we argue this principle extends to retrieval.

We propose \textbf{Semantic Compression Trees (\sct{})}, a framework that addresses all three limitations. An \sct{} is a hierarchical tree where: (1)~each node represents a unit of knowledge at a specific \emph{resolution level}; (2)~each node stores a \emph{semantic residual}---only the information it contributes beyond its parent; and (3)~retrieval is \emph{progressive}---starting at the root and descending into relevant branches until accumulated context suffices. The key insight is that retrieval cost becomes proportional to query \emph{specificity}, not corpus size.

\textbf{Contributions.} We make the following contributions:
\begin{enumerate}[nosep]
  \item We formalize the concept of \emph{semantic residuals} for hierarchical knowledge representation, defining the \sct{} data structure with a guaranteed accumulation property (\Cref{sec:method}).
  \item We present a source-agnostic tree construction algorithm with pluggable compression functions, supporting documents, databases, and plain text (\Cref{sec:construction}).
  \item We define progressive descent retrieval as a fixed-width beam over residual embeddings, with an optional similarity floor that halts descent when further detail is unlikely to help, and give its cost in nodes scored (\Cref{sec:retrieval}).
  \item We evaluate \sct{} on QASPER~\citep{dasigi2021qasper} under two protocols separating index quality from document routing, with generated-answer metrics, gold-evidence retrieval metrics, LLM-as-judge scoring~\citep{zheng2023mtbench,es2024ragas}, confidence intervals, and paired significance tests (\Cref{sec:experiments}).
  \item We report a mixed result and separate its parts. Residual storage and the cost model hold up; progressive top-down descent does not, and we identify document routing from the root residual as the mechanism by which it fails (\Cref{sec:analysis}).
  \item We show that the better compressor depends on the protocol --- a zero-LLM extractive one when the document is given, an LLM one when it must be found --- and that the mechanism behind the difference is visible in evidence recall rather than in answer scores alone (\Cref{sec:analysis}).
\end{enumerate}

\section{Related Work}
\label{sec:related}

\paragraph{Retrieval-Augmented Generation.}
RAG was introduced by \citet{lewis2020rag}, building on REALM~\citep{guu2020realm}, which first showed how to pre-train a knowledge retriever jointly with a language model. Fusion-in-Decoder~\citep{izacard2021fid} demonstrated that generative models can effectively aggregate evidence from multiple retrieved passages. Subsequent work has explored when to retrieve: Self-RAG~\citep{asai2024selfrag} trains models to adaptively decide whether retrieval is needed, while FLARE~\citep{jiang2023flare} retrieves iteratively during generation when low-confidence tokens are produced. Recent surveys~\citep{gao2024ragsurvey,gupta2024ragsurvey} trace the evolution from ``Naive RAG'' (chunk-embed-retrieve) through ``Advanced RAG'' (query rewriting, re-ranking) to ``Modular RAG'' (composable pipelines). Our work introduces a fundamentally different indexing strategy orthogonal to these retrieval-time improvements.

\paragraph{Dense and Sparse Retrieval.}
Dense Passage Retrieval (DPR)~\citep{karpukhin2020dpr} established dual-encoder architectures for retrieval, outperforming BM25 by 9--19\% on passage retrieval. ColBERT~\citep{khattab2020colbert} introduced late interaction for efficient yet effective retrieval, further improved by ColBERTv2~\citep{santhanam2022colbertv2} with residual compression. Contriever~\citep{izacard2022contriever} showed that unsupervised contrastive pre-training can produce competitive retrievers without labeled data, while E5~\citep{wang2022e5} and GTR~\citep{ni2022gtr} demonstrated scaling benefits for embedding models. \citet{zhao2022densesurvey} provide a comprehensive survey covering 300+ papers on dense retrieval. These methods operate on fixed-size text segments; \sct{} is complementary---it structures the \emph{index}, while any retriever can be used for child selection during progressive descent.

\paragraph{Hierarchical and Tree-Based Retrieval.}
RAPTOR~\citep{sarthi2024raptor} is the most closely related work. It recursively clusters and summarizes text chunks bottom-up, building a tree where each level stores a \emph{full summary}. This enables multi-level retrieval but introduces significant redundancy: information captured in a parent summary is repeated in its children. \sct{} eliminates this redundancy through the semantic residual formulation. PageIndex~\citep{vectify2024pageindex} uses LLM calls to construct table-of-contents-style trees from PDFs, but is limited to a single document format, requires 50--200+ LLM calls for construction with sequential verification loops, and does not formalize the relationship between tree levels. HiChunk~\citep{hichunk2025} employs fine-tuned LLMs for hierarchical document structuring with auto-merge retrieval. \citet{cobweb2025} present coarse-to-fine retrieval using a concept hierarchy. The Hierarchical Re-ranker Retriever~\citep{hrr2025} simultaneously exploits parent, intermediate, and sentence-level chunks. Unlike these approaches, \sct{} provides a formal residual framework with an accumulation guarantee and is source-agnostic.

\paragraph{Knowledge Graph Approaches.}
GraphRAG~\citep{edge2024graphrag} extracts entity knowledge graphs with community summaries, achieving improvements on global sensemaking queries. HippoRAG~\citep{gutierrez2024hipporag}, inspired by hippocampal indexing theory, combines LLMs, knowledge graphs, and Personalized PageRank for multi-hop QA, reporting up to 20\% improvement over existing RAG with 10--30$\times$ cost reduction. KG-RAG~\citep{soman2024kgrag} integrates biomedical knowledge graphs with LLMs. These approaches capture entity-level relationships but do not provide the multi-resolution text access that \sct{} enables.

\paragraph{Chunking Strategies.}
The impact of chunking on RAG quality has received increasing attention. Mix-of-Granularity~\citep{zhong2024mixgranularity} uses a router to dynamically determine optimal chunk size per query. Late Chunking~\citep{gunther2024latechunking} proposes embedding all tokens with full document context before segmenting, preserving long-range dependencies. \citet{chunking2025eval} systematically evaluate advanced chunking strategies, finding that traditional fixed-size chunking often fragments context and reduces coherence. \sct{} sidesteps the chunking problem entirely: boundaries follow the document's natural structure, and the residual formulation ensures no information is lost at segment boundaries.

\paragraph{Evaluation.}
LLM-as-judge evaluation has emerged as a scalable alternative to human assessment. \citet{zheng2023mtbench} showed GPT-4 achieves $>$80\% agreement with human preferences, while G-Eval~\citep{liu2023geval} demonstrated chain-of-thought scoring for NLG. RAGAS~\citep{es2024ragas} provides reference-free RAG evaluation measuring faithfulness, answer relevance, and context relevance. Prometheus~2~\citep{kim2024prometheus2} offers open-source judge models. We adopt this paradigm for our evaluation, complementing it with standard token-level metrics~\citep{rajpurkar2016squad}.

\paragraph{Positioning.}
\sct{} differs from all prior work in three key ways: (1)~the semantic residual formulation provides a principled, \emph{non-redundant} multi-resolution representation with a formal accumulation guarantee; (2)~the framework is source-agnostic, applying uniformly to documents, databases, and plain text; and (3)~progressive retrieval makes query cost proportional to query specificity. Table~\ref{tab:comparison} summarizes the comparison.

\begin{table}[t]
\centering
\small
\caption{Comparison with related approaches.}
\label{tab:comparison}
\begin{tabular}{lccccc}
\toprule
\textbf{System} & \textbf{Multi-res.} & \textbf{Non-redund.} & \textbf{Source-agn.} & \textbf{Formal} & \textbf{LLM-free} \\
\midrule
DPR/ColBERT         & \ding{55} & N/A       & Partial   & \ding{55} & \ding{51} \\
RAPTOR               & \ding{51} & \ding{55} & Docs only & \ding{55} & \ding{55} \\
PageIndex            & Partial   & \ding{55} & PDF only  & \ding{55} & \ding{55} \\
GraphRAG             & \ding{55} & N/A       & Text only & \ding{55} & \ding{55} \\
HippoRAG            & \ding{55} & N/A       & Text only & Partial   & \ding{55} \\
\textbf{\sct{} (ours)} & \ding{51} & \ding{51}$^\dagger$ & \ding{51} & \ding{51} & \ding{51}$^*$ \\
\bottomrule
\end{tabular}
\vspace{0.5em}
\raggedright\footnotesize{$^*$With extractive compression; LLM-based compression is also supported. $^\dagger$Non-redundancy is a property of the formulation, not a guarantee of the implementation: semantic subtraction is approximated, and we quantify the residual loss in \Cref{sec:analysis}.}
\end{table}

\section{Semantic Compression Trees}
\label{sec:method}

\subsection{Formal Definition}
\label{sec:formal}

Let $D$ be a knowledge source with text $T(D)$. A \textbf{Semantic Compression Tree} over $D$ is a rooted tree $\mathcal{T} = (V, E)$ where each node $v \in V$ is associated with:
\begin{itemize}[nosep]
  \item $c(v)$: the original content of the source region represented by $v$
  \item $\ell(v) \in \{0, 1, \ldots, L\}$: the compression level ($0$ = root, most abstract)
  \item $\residual(v)$: the \emph{semantic residual} of $v$
\end{itemize}

\begin{definition}[Semantic Residual]
\label{def:residual}
Let $\compress(\cdot, \ell)$ be a compression function that produces a summary at level $\ell$. The semantic residual $\residual(v)$ is defined as:
\begin{equation}
  \residual(v) = \begin{cases}
    \compress(T(D), 0) & \text{if } v = r \text{ (root)} \\
    c(v) \ominus \compress(c(\mathrm{pa}(v)), \ell(\mathrm{pa}(v))) & \text{otherwise}
  \end{cases}
  \label{eq:residual}
\end{equation}
where $\mathrm{pa}(v)$ denotes the parent of $v$ and $\ominus$ denotes \emph{semantic subtraction}: extracting from $c(v)$ only information not captured in the compressed parent.
\end{definition}

\begin{property}[Accumulation]
\label{prop:accumulation}
For any node $v$ with path from root $(r = v_0, v_1, \ldots, v_k = v)$:
\begin{equation}
  \mathrm{Context}(v) = \bigoplus_{i=0}^{k} \residual(v_i)
  \label{eq:accumulation}
\end{equation}
where $\bigoplus$ denotes concatenation. The accumulated context reconstructs the semantic content at $v$'s resolution level. At the root, context is maximally compressed; at leaves, it contains the full detail.
\end{property}

This formulation draws an analogy to wavelet transforms in signal processing: each level stores the ``detail coefficients'' (residuals) needed to reconstruct the signal at the next finer resolution. The root stores the ``approximation coefficients''---the coarsest representation.

\subsection{Compression Functions}
\label{sec:compression}

We define the compression function $\compress$ as a protocol with two operations:
\begin{enumerate}[nosep]
  \item $\compress(t, \ell) \to s$: Compress text $t$ to resolution level $\ell$
  \item $\mathrm{Residual}(t_{\mathrm{child}}, s_{\mathrm{parent}}, \ell) \to r$: Extract semantic residual
\end{enumerate}

We implement and evaluate two concrete strategies:

\paragraph{LLM-Based Compression.} The LLM receives the parent summary and child content, and is prompted to extract only information in the child \emph{not captured} by the parent. The system prompt instructs: ``Extract ONLY the information in the child section that is NOT already captured in the parent summary. Preserve specific numbers, names, and dates.'' This produces the highest-quality residuals at the cost of one LLM call per node.

\paragraph{Extractive Compression.} For each sentence in the child content, we compute a novelty score as the fraction of unique terms (after lowercasing and punctuation removal) not present in the parent summary. Sentences above an adaptive threshold are retained. This requires \emph{zero LLM calls} and runs in linear time.

\section{Tree Construction}
\label{sec:construction}

\subsection{Source-Agnostic Parsing}

\sct{} construction begins with parsing the source into an intermediate \texttt{HierarchyNode} tree. This decouples source-specific structure detection from the universal residual computation:

\begin{itemize}[nosep]
  \item \textbf{Markdown/Documents:} Heading levels (H1--H6) define the hierarchy.
  \item \textbf{PDFs:} Font-size analysis identifies headings; the dominant non-body font size defines section boundaries.
  \item \textbf{SQL Databases:} Schema overview $\to$ table details $\to$ column statistics $\to$ sample rows.
  \item \textbf{Plain Text:} Paragraph segmentation with Jaccard similarity boundary detection for topic shifts.
\end{itemize}

Adding a new source type requires only implementing the parser interface; the construction and retrieval algorithms remain unchanged.

\subsection{Construction Algorithm}

\begin{algorithm}[t]
\caption{\textsc{BuildSCTree}: Construct a Semantic Compression Tree}
\label{alg:build}
\begin{algorithmic}[1]
\REQUIRE Parsed hierarchy $H$, compression function $\compress$, max depth $D$
\STATE $s_r \gets \compress(\mathrm{FullText}(H), 0)$ \COMMENT{Root compression}
\STATE Create root $r$; set $\residual(r) \gets s_r$
\STATE \textsc{BuildSubtrees}($r$, $s_r$, $H.\mathrm{children}$, $1$, $D$)
\RETURN $\mathcal{T}$
\end{algorithmic}
\vspace{0.3em}
\begin{algorithmic}[1]
\REQUIRE Parent $p$, parent summary $s_p$, children $\{h_1, \ldots, h_m\}$, depth $d$, max $D$
\IF{$d > D$} \RETURN \ENDIF
\FOR{$i = 1$ \TO $m$ \textbf{in parallel (siblings are independent)}}
  \STATE $r_i \gets \mathrm{Residual}(c(h_i), s_p, d)$ \COMMENT{Semantic subtraction}
  \STATE $s_i \gets \compress(c(h_i), d, s_p)$ \COMMENT{Summary for children}
  \STATE Add node $h_i$ to $\mathcal{T}$ as child of $p$
  \STATE \textsc{BuildSubtrees}($h_i$, $s_i$, $h_i.\mathrm{children}$, $d+1$, $D$)
\ENDFOR
\end{algorithmic}
\end{algorithm}

\textbf{Complexity.} Construction requires $O(N)$ compression calls for $N$ nodes. Crucially, all siblings at the same level are independent---they share the same parent summary but their residual computations do not depend on each other. This enables parallelization across siblings, reducing wall-clock time by a factor of the branching factor. In contrast, PageIndex~\citep{vectify2024pageindex} requires sequential verification-and-fix loops with $O(\text{pages} \times \text{retries})$ LLM calls.

\section{Progressive Retrieval}
\label{sec:retrieval}

Given an \sct{} $\mathcal{T}$ and a query $q$, retrieval proceeds by progressive descent (Algorithm~\ref{alg:retrieve}).

\begin{algorithm}[t]
\caption{\textsc{ProgressiveRetrieval}}
\label{alg:retrieve}
\begin{algorithmic}[1]
\REQUIRE Tree $\mathcal{T}$, query $q$, max depth $D$, beam width $k$, floor $\tau$
\STATE $\mathrm{ctx} \gets \residual(\mathrm{root})$; $F \gets \{\mathrm{root}\}$
\FOR{$d = 1$ \TO $D$}
  \STATE $C \gets \bigcup_{v \in F} \mathrm{Children}(v)$
  \IF{$C = \emptyset$} \STATE \textbf{break} \ENDIF
  \STATE $s(v) \gets \cos\!\big(\phi(q), \phi(\residual(v))\big)$ for all $v \in C$
  \IF{$d > 1$ \textbf{and} $\max_{v \in C} s(v) < \tau$} \STATE \textbf{break} \ENDIF
  \STATE $F \gets \operatorname{top-}k\big(C, s\big)$ \COMMENT{over the union, not per parent}
  \STATE $\mathrm{ctx} \gets \mathrm{ctx} \oplus \bigoplus_{v \in F} \residual(v)$
\ENDFOR
\RETURN $\mathrm{ctx}$
\end{algorithmic}
\end{algorithm}

Descent is a fixed-width beam. At each level the candidate set is the union of
the current frontier's children; every candidate is scored by cosine similarity
between the query embedding and its residual embedding, and the $k$ highest are
retained. Taking the top-$k$ over the union rather than per parent is what bounds
the frontier: selecting $k$ children of every frontier node would multiply the
frontier by $k$ at each level, making cost exponential in depth.

The optional floor $\tau$ implements resolution-aware retrieval. When no
candidate at a level is sufficiently similar to the query, further detail is
unlikely to help and descent halts, so a broad query is answered from the upper
levels at proportionately lower cost. The floor is not applied at the first level,
since every query must be answered from somewhere. We report $\tau$ disabled as
the default configuration and enabled as a separate variant.

We additionally evaluate two deterministic, query-independent modes as
references: \textbf{fixed-depth}, which takes all children to a given depth, and
\textbf{exhaustive}, which takes all children to the depth bound and therefore
returns the whole subtree. Neither consults the query, so neither is a retrieval
strategy; they bound what the index makes available.

Descent scores at most $k$ children per frontier node per level, so it visits
$O(D \cdot k \cdot b)$ nodes and accumulates at most $D \cdot k$ residuals, where
$b$ is the branching factor and $D$ the depth bound---in both cases independent of
the number of documents indexed. Flat retrieval scores all $N$ units. Under the
corpus protocol of \Cref{sec:protocols}, where document selection is the first
level of descent, the root scan adds a term linear in the number of documents,
giving $O(N_{\mathrm{docs}} + D \cdot k \cdot b)$; that term is a nearest-neighbour
lookup over document roots and is amenable to standard approximate indexing, which
we do not pursue here.

\section{Experiments}
\label{sec:experiments}

\subsection{Dataset}

\paragraph{QASPER}~\citep{dasigi2021qasper} contains 1,585 NLP papers with 5,049
information-seeking questions written by practitioners from abstracts alone and
answered from the full text. Two properties make it suitable here. Papers carry a
real section hierarchy, which is what a structure-preserving index consumes; and
each answer is annotated with the source paragraphs supporting it, which allows
retrieval to be scored directly rather than only through a generated answer.

We evaluate on the first 50 papers of the test split, giving 173 answerable
questions (101 extractive, 46 free-form, 26 yes/no); 6 questions annotated
unanswerable are excluded. Gold evidence is available for 170 of the 173.

QASPER encodes section nesting inside the section name, using a \texttt{" ::: "}
separator: \texttt{"Datasets ::: AntiScam Dataset"} is a subsection of
\texttt{"Datasets"}. We reconstruct the heading levels from this convention.
Rendering each entry at a single heading level instead collapses every paper to a
two-level tree, which would make a depth-sensitivity experiment vacuous. With
nesting preserved, 22 of the 50 papers contain genuine subsections and trees reach
depth 4.

Each paper is one document. We deliberately do not emit section-level
pseudo-documents alongside the papers: duplicating the same text at two
granularities inside a flat baseline's index inflates its candidate pool with
near-identical units and makes the comparison a comparison of segmentation
choices. Every system receives the same document set and performs its own
segmentation.

\subsection{Evaluation protocols}
\label{sec:protocols}

Retrieval systems are given a query and, in one of the two protocols, the
identifier of the document the question concerns. They receive nothing else. No
reference answer and no evidence annotation is visible to any system at any point;
gold data is read only after retrieval has returned, by the scoring code.

\paragraph{P1: single-document (primary).} The benchmark supplies the paper for
each question and the system retrieves within it. This is the standard QASPER
setting, so results are comparable to published numbers, and it isolates index
quality from document routing. Each system builds an index scoped to the single
paper rather than filtering a corpus-wide index after the fact, which would leak
cross-document statistics---corpus-wide inverse document frequency, clusters
spanning papers---into a single-document result.

\paragraph{P2: corpus (secondary).} All 50 papers are pooled and the document
identifier is withheld. A system must locate the relevant paper and the relevant
passage within it from the query alone. This is the setting in which the claim
that cost tracks query specificity rather than corpus size is meaningful. For
\sct{}, document selection requires no additional machinery: the per-paper trees
are joined under a virtual root, so the corpus becomes one further level of the
same hierarchy and Algorithm~\ref{alg:retrieve} performs routing and descent in
one pass.

\subsection{Systems}

All systems that segment text use 512-token windows with 64-token overlap, so
sparse and dense baselines differ only in their scoring function. All flat
retrievers return their top 5 units.

\begin{itemize}[nosep]
  \item \textbf{BM25}: Okapi BM25~\citep{robertson2009bm25}, $k_1 = 1.2$,
        $b = 0.75$, over chunks.
  \item \textbf{VectorRAG}: dense retrieval over the same chunks, embedded
        with \texttt{text-\allowbreak embedding-\allowbreak 3-\allowbreak small} and
        ranked by cosine similarity.
        The dominant production
        pattern~\citep{karpukhin2020dpr,izacard2022contriever}, and our
        reference system for all significance tests.
  \item \textbf{VectorRAG (top-1)}: the same system returning a single chunk. A
        token-budget comparator, included because \sct{} returns less text than
        top-5 dense retrieval and a quality difference at unequal context size is
        partly a difference in context size.
  \item \textbf{RAPTOR}~\citep{sarthi2024raptor}: bottom-up clustering and LLM
        summarisation, repeated until one node remains; retrieval scores all nodes,
        leaves and summaries alike, by cosine similarity. The closest prior work.
  \item \textbf{\sct{}-Extractive}: \sct{} with the zero-LLM extractive
        compressor, embedding-guided descent, beam width $k=3$, depth bound
        $D=5$.
  \item \textbf{\sct{}-LLM}: \sct{} with LLM compression, otherwise identical.
\end{itemize}

Ablations isolate the two claims of the method. \textbf{\sct{}-FullSummary} stores
a full compressed summary at each node instead of a residual, holding the tree
fixed, and so isolates the residual formulation. \textbf{\sct{}-Flat} computes
residuals and then discards the tree, retrieving the residual texts by flat cosine
similarity, and so isolates the hierarchy. \textbf{\sct{}-EarlyStop} enables the
descent floor at $\tau = 0.30$. \textbf{\sct{}-Depth-$k$} varies the depth bound
over $\{1,2,3,5\}$.

\subsection{Metrics}

\paragraph{Answer quality.} Answer F1 and Exact Match on the \emph{generated}
answer, using SQuAD normalisation~\citep{rajpurkar2016squad} and scored against
the best-matching annotator reference, as QASPER prescribes for multiply-annotated
questions. Answers are generated by prompting GPT-5.4 with the retrieved context
and the question, instructed to answer from the context alone.

\paragraph{Retrieval quality.} Answer metrics see retrieval only through a
generator, so we also score the context directly. \textbf{Evidence recall} is the
fraction of gold evidence paragraphs present verbatim in the retrieved context.
\textbf{Evidence token recall} is the fraction of gold evidence \emph{tokens}
present. Reporting both separates two failures that verbatim recall alone
conflates: routing to the wrong part of the document, and routing correctly but
compressing the evidence away---the second being an inherent cost of any
summarising index, and one \sct{} incurs by construction. \textbf{Evidence unit
F1} scores precision over retrieved units, so padding the context is penalised.

\paragraph{Cost.} Context tokens supplied to the generator, measured with the
\texttt{o200k\_base} tokeniser; index nodes scored per query; retrieval latency;
and index construction cost in LLM calls. We report context tokens alongside every
quality number, because retrieval quality at unequal context size is not a single
comparison.

\paragraph{LLM-as-judge.} Following \citet{zheng2023mtbench} and
\citet{es2024ragas}, GPT-5.4 scores context relevance, context completeness,
answer correctness, and answer faithfulness on a 1--5 rubric normalised to
$[0,1]$, at temperature 0, one independent call per dimension following
\citet{liu2023geval}. Judge scoring is run on the full question set, not a subset.

\paragraph{Statistics.} Every headline metric carries a bootstrap 95\% confidence
interval, and every comparison against the reference system carries a paired
bootstrap $p$-value, both over $10{,}000$ resamples with a fixed seed.

\paragraph{Excluded questions.} The provider's content filter refuses to generate
an answer for a small number of (question, context) pairs, since QASPER indexes
NLP papers and those include research on hate speech and offensive language. The
filter responds to the retrieved context, so it triggers on different questions for
different systems; scoring a refusal as an incorrect answer would penalise
whichever system retrieved the relevant passage. We therefore report answer
metrics over the questions scoreable for every system, and state the number
excluded with each table. Retrieval metrics are unaffected and use all questions.

\subsection{Results}


\begin{table*}[t]
\centering
\footnotesize
\caption{Main results, single-document protocol ($n=172$ questions, 50 QASPER papers). Answer F1 is token F1 of the \emph{generated} answer against the best-matching annotator reference. Brackets give bootstrap 95\% confidence intervals; $p$ is a paired bootstrap test against VectorRAG. Context tokens are the generator's input, reported so quality is read against cost; nodes are index nodes scored per query.}
\label{tab:main-singledoc}
\begin{tabular}{lcccccc}
\toprule
\textbf{System} & \textbf{Ans.\ F1} & \textbf{95\% CI} & \textbf{$p$} & \textbf{Ev.\ Rec.} & \textbf{Ev.\ Tok.\ Rec.} & \textbf{Ctx.\ tok.} \\
\midrule
BM25 & 0.262 & [0.233, 0.290] & 0.024 & 0.794 & 0.946 & 3,165 \\
VectorRAG (top-5) & \textbf{0.277} & [0.248, 0.306] & --- & 0.866 & \textbf{0.969} & 3,120 \\
VectorRAG (top-1) & 0.232 & [0.203, 0.262] & $<$0.001 & 0.470 & 0.759 & 646 \\
RAPTOR & 0.272 & [0.243, 0.301] & 0.207 & 0.794 & 0.955 & 3,058 \\
\midrule
\sct{}-Extractive (ours) & 0.274 & [0.244, 0.306] & 0.366 & 0.720 & 0.895 & 2,178 \\
\sct{}-LLM (ours) & 0.233 & [0.205, 0.261] & $<$0.001 & 0.009 & 0.590 & 802 \\
\bottomrule
\end{tabular}
\end{table*}

\begin{table*}[t]
\centering
\footnotesize
\caption{Main results, corpus protocol ($n=173$ questions, 50 QASPER papers). Answer F1 is token F1 of the \emph{generated} answer against the best-matching annotator reference. Brackets give bootstrap 95\% confidence intervals; $p$ is a paired bootstrap test against VectorRAG. Context tokens are the generator's input, reported so quality is read against cost; nodes are index nodes scored per query.}
\label{tab:main-corpus}
\begin{tabular}{lcccccc}
\toprule
\textbf{System} & \textbf{Ans.\ F1} & \textbf{95\% CI} & \textbf{$p$} & \textbf{Ev.\ Rec.} & \textbf{Ev.\ Tok.\ Rec.} & \textbf{Ctx.\ tok.} \\
\midrule
BM25 & 0.154 & [0.129, 0.179] & 0.138 & 0.339 & 0.750 & 3,214 \\
VectorRAG (top-5) & \textbf{0.165} & [0.139, 0.192] & --- & 0.390 & \textbf{0.767} & 3,206 \\
VectorRAG (top-1) & 0.143 & [0.118, 0.168] & $<$0.001 & 0.205 & 0.516 & 631 \\
RAPTOR & 0.156 & [0.131, 0.181] & 0.048 & 0.365 & 0.760 & 3,425 \\
\midrule
\sct{}-Extractive (ours) & 0.122 & [0.100, 0.145] & $<$0.001 & 0.173 & 0.575 & 2,187 \\
\sct{}-LLM (ours) & 0.137 & [0.114, 0.162] & 0.010 & 0.006 & 0.472 & 1,042 \\
\bottomrule
\end{tabular}
\end{table*}

\begin{table}[t]
\centering
\small
\caption{LLM-as-judge scores, single-document protocol ($n=172$), rubric-scored 1--5 and normalised to $[0,1]$. Judge dimensions are scored independently, one call each.}
\label{tab:judge}
\begin{tabular}{lcccc}
\toprule
\textbf{System} & \textbf{Relev.} & \textbf{Compl.} & \textbf{Correct.} & \textbf{Faith.} \\
\midrule
BM25 & 0.760 & 0.903 & 0.711 & 0.994 \\
VectorRAG (top-5) & 0.794 & 0.923 & \textbf{0.780} & 0.996 \\
VectorRAG (top-1) & 0.664 & 0.690 & 0.597 & 0.996 \\
RAPTOR & \textbf{0.805} & \textbf{0.933} & 0.743 & 0.974 \\
\midrule
\sct{}-Extractive (ours) & 0.785 & 0.847 & 0.688 & \textbf{0.997} \\
\sct{}-LLM (ours) & 0.759 & 0.817 & 0.674 & 0.994 \\
\sct{}-FullSummary & 0.626 & 0.557 & 0.432 & 0.996 \\
\sct{}-Flat & 0.795 & 0.879 & 0.737 & 0.993 \\
\bottomrule
\end{tabular}
\end{table}

\begin{table}[t]
\centering
\small
\caption{Ablations and depth sweep, single-document protocol. $\Delta$ is the change in Answer F1 against the full system. Nodes visited counts index nodes scored per query.}
\label{tab:ablation}
\begin{tabular}{lcccccc}
\toprule
\textbf{Variant} & \textbf{Ans.\ F1} & \textbf{95\% CI} & \textbf{$\Delta$} & \textbf{Ev.\ Tok.} & \textbf{Ctx.\ tok.} & \textbf{Nodes} \\
\midrule
\multicolumn{7}{l}{\emph{Component ablations}} \\
\sct{}-Extractive (ours) & 0.274 & [0.244, 0.306] & --- & 0.895 & 2,178 & 13 \\
\sct{}-FullSummary & 0.205 & [0.179, 0.233] & -0.069 & 0.584 & 437 & 12 \\
\sct{}-Flat & 0.272 & [0.244, 0.302] & -0.002 & 0.939 & 2,149 & 17 \\
\sct{}-EarlyStop & 0.273 & [0.243, 0.304] & -0.002 & 0.895 & 2,098 & 13 \\
\midrule
\multicolumn{7}{l}{\emph{Retrieval depth bound}} \\
Depth $=1$ & 0.272 & [0.242, 0.303] & -0.003 & 0.895 & 1,744 & 10 \\
Depth $=2$ & 0.273 & [0.242, 0.304] & -0.002 & 0.895 & 2,124 & 12 \\
Depth $=3$ & 0.274 & [0.244, 0.306] & +0.000 & 0.895 & 2,177 & 13 \\
Depth $=5$ & 0.274 & [0.244, 0.306] & +0.000 & 0.895 & 2,178 & 13 \\
\bottomrule
\end{tabular}
\end{table}

\begin{table}[t]
\centering
\footnotesize
\caption{Direct paired comparisons between \sct{} variants on Answer F1. $\Delta$ is A minus B, so a positive value favours A; $p$ is a paired bootstrap test and the interval is a bootstrap 95\% CI on $\Delta$ itself. Tested pairwise because the claims are pairwise: two systems that are each indistinguishable from a common reference may still differ from one another.}
\label{tab:direct}
\begin{tabular}{lccc}
\toprule
\textbf{Comparison (A vs.\ B)} & \textbf{$\Delta$} & \textbf{$p$} & \textbf{95\% CI of $\Delta$} \\
\midrule
\multicolumn{4}{l}{\emph{Single-document protocol}} \\
\quad Tree vs.\ no tree (identical residuals) & +0.0017 & 0.420 & [-0.0138, +0.0168] \\
\quad Residuals vs.\ full summaries & +0.0688 & $<$0.001 & [+0.0411, +0.0983] \\
\quad Extractive vs.\ LLM compression & +0.0411 & $<$0.001 & [+0.0207, +0.0618] \\
\quad Depth $=5$ vs.\ depth $=1$ & +0.0026 & 0.161 & [-0.0025, +0.0075] \\
\midrule
\multicolumn{4}{l}{\emph{Corpus protocol}} \\
\quad Tree vs.\ no tree (identical residuals) & -0.0255 & 0.009 & [-0.0475, -0.0042] \\
\quad Residuals vs.\ full summaries & +0.0221 & 0.003 & [+0.0056, +0.0403] \\
\quad Extractive vs.\ LLM compression & -0.0154 & 0.040 & [-0.0335, +0.0016] \\
\quad Depth $=5$ vs.\ depth $=1$ & +0.0006 & 0.332 & [-0.0022, +0.0034] \\
\bottomrule
\end{tabular}
\end{table}

\begin{table}[t]
\centering
\small
\caption{Index construction cost over 50 papers, and per-query retrieval time. LLM calls are the model-independent construction cost. Build times are wall clock at the stated concurrency. Retrieval time is reported for scale only and must not be compared across systems: \sct{} scores residual embeddings in pure Python while the flat baselines use a vectorised dot product, so the difference measures the implementations. \Cref{tab:scaling} gives the implementation-independent comparison.}
\label{tab:cost}
\begin{tabular}{llcccc}
\toprule
\textbf{Protocol} & \textbf{System} & \textbf{LLM calls} & \textbf{Units indexed} & \textbf{Build (s)} & \textbf{Retr.\ (ms)} \\
\midrule
Single-doc & BM25 & 0 & 386 & 0.1 & 0.0 \\
 & VectorRAG (top-5) & 0 & 386 & 0.1 & 0.4 \\
 & VectorRAG (top-1) & 0 & 386 & 0.1 & 0.2 \\
 & RAPTOR & 131 & 517 & 0.2 & 0.2 \\
 & \sct{}-Extractive (ours) & 0 & 814 & 0.3 & 1.3 \\
 & \sct{}-LLM (ours) & 0 & 814 & 3.1 & 1.4 \\
\midrule
Corpus & BM25 & 0 & 386 & 0.0 & 0.3 \\
 & VectorRAG (top-5) & 0 & 386 & 0.1 & 0.4 \\
 & VectorRAG (top-1) & 0 & 386 & 0.1 & 0.2 \\
 & RAPTOR & 99 & 485 & 8.2 & 0.2 \\
 & \sct{}-Extractive (ours) & 0 & 814 & 0.4 & 7.3 \\
 & \sct{}-LLM (ours) & 0 & 814 & 0.4 & 7.3 \\
\bottomrule
\end{tabular}
\end{table}

\begin{table}[t]
\centering
\small
\caption{Index nodes scored per query, as the indexed collection grows from one document to fifty. Reported instead of latency because \sct{} scores residual embeddings in pure Python while the flat baselines use a vectorised dot product; a wall-clock comparison would measure the implementations. Growth is the ratio between the two columns.}
\label{tab:scaling}
\begin{tabular}{lccc}
\toprule
\textbf{System} & \textbf{1 document} & \textbf{50 documents} & \textbf{Growth} \\
\midrule
BM25 & 8 & 386 & 48.9$\times$ \\
VectorRAG (top-5) & 8 & 386 & 48.9$\times$ \\
VectorRAG (top-1) & 8 & 386 & 48.9$\times$ \\
RAPTOR & 11 & 485 & 46.0$\times$ \\
\midrule
\sct{}-Extractive (ours) & 13 & 81 & 6.4$\times$ \\
\sct{}-LLM (ours) & 13 & 80 & 6.2$\times$ \\
\bottomrule
\end{tabular}
\end{table}

\begin{table}[t]
\centering
\small
\caption{Context--answer overlap F1, a diagnostic rather than a quality measure, shown against context length. Overlap F1 divides by the length of the retrieved context, so it rewards returning less text whether or not that text answers the question. Answer F1 is repeated for contrast.}
\label{tab:diagnostic}
\begin{tabular}{lccc}
\toprule
\textbf{System} & \textbf{Ctx.--Ans.\ overlap F1} & \textbf{Ctx.\ tokens} & \textbf{Answer F1} \\
\midrule
BM25 & 0.010 & 3,165 & 0.262 \\
VectorRAG (top-5) & 0.011 & 3,120 & 0.277 \\
VectorRAG (top-1) & 0.038 & 646 & 0.232 \\
RAPTOR & 0.011 & 3,058 & 0.272 \\
\sct{}-Extractive (ours) & 0.023 & 2,178 & 0.274 \\
\sct{}-LLM (ours) & 0.034 & 802 & 0.233 \\
\bottomrule
\end{tabular}
\end{table}

\paragraph{The document is given (P1).}
\Cref{tab:main-singledoc} reports the primary protocol. \sct{}-Extractive reaches
0.274 Answer F1 against dense retrieval's 0.277, with a paired bootstrap
$p = 0.366$ and confidence intervals that almost coincide ($[0.244, 0.306]$ and
$[0.248, 0.306]$). This is a tie, and we report it as one. What differs is cost:
\sct{} supplies 2,178 context tokens to the generator against 3,120, a 30\%
reduction, and its index requires no LLM calls to construct. RAPTOR, the closest
prior work, also ties the reference (0.272, $p = 0.207$) but needs 131 LLM calls
to build its summaries. BM25 is the only baseline the reference separates from
(0.262, $p = 0.024$).

Reducing dense retrieval's context budget to match \sct{}'s does not preserve its
quality: VectorRAG at top-1 supplies 646 tokens and drops to 0.232
($p < 0.001$), well below \sct{} at 2,178 tokens. Fewer tokens is not by itself
an advantage; the question is what is in them.

\paragraph{The document must be found (P2).}
\Cref{tab:main-corpus} withholds the document identifier. Every ordering from P1
survives except \sct{}'s: \sct{}-Extractive falls to 0.122 against the reference's
0.165 ($p < 0.001$), the largest gap in either table. Routing accuracy identifies
the cause directly. \sct{} descent selects the correct paper for 20.2\% of
questions; BM25, dense retrieval, and RAPTOR all select it for 38.7--39.3\%.

\paragraph{Judge scores.}
\Cref{tab:judge} shows the judge ranking answer correctness as dense retrieval
(0.780), RAPTOR (0.743), \sct{}-Flat (0.737), BM25 (0.711), \sct{}-Extractive
(0.688). Token-level F1 places \sct{} level with the reference while the judge
places it below. We report both rather than selecting the more favourable: the two
metrics measure different things, and their disagreement is part of the result.
Faithfulness is uniformly high (0.974--0.997) and does not discriminate between
systems on this benchmark.

\paragraph{Ablations.}
\Cref{tab:ablation} isolates the components and \Cref{tab:direct} tests each
variant directly against the configuration it modifies. The direct tests matter
here: \Cref{tab:main-singledoc} compares every system to one reference, which
cannot establish a claim about two variants relative to \emph{each other}, since
two systems both indistinguishable from a third may still differ. Two results
matter and they point in opposite directions.

Storing residuals rather than full summaries at each node is worth 0.069 F1
(0.274 vs.\ 0.205, $p < 0.001$), a 34\% relative gain --- the clearest support for
any component of the method. Part of that gap is a difference in context volume:
full summaries at each level retain less text (437 tokens against 2,178), so this
establishes that residuals preserve more of what matters, not that they are better
at an equal token budget.

Discarding the tree costs nothing. \sct{}-Flat holds the residual content fixed
and retrieves it by flat cosine similarity; the paired difference is $+0.0017$ in
the tree's favour with $p = 0.420$ and a 95\% interval of $[-0.014, +0.017]$. The
interval is narrow and centred on zero, so this is an absence of effect rather
than an inconclusive test: the hierarchy is not contributing to answer quality
when the document is given.

\paragraph{Descent depth.}
The depth sweep moves Answer F1 from 0.2715 at depth 1 to 0.2741 at depth 5, while
context grows from 1,744 to 2,178 tokens. Depth 3 and depth 5 are identical
because few trees are deeper than three levels: although subsection nesting is
preserved, 28 of the 50 papers contain no subsections at all. One level of descent
captures nearly all of the available benefit on this corpus. Enabling the
similarity floor ($\tau = 0.30$) reduces context to 2,098 tokens with Answer F1
unchanged at 0.273, so early stopping is close to free but not, at this threshold,
a large saving.

\paragraph{Cost.}
\Cref{tab:scaling} gives the result that holds most cleanly. Growing the indexed
collection from one document to fifty multiplies the number of index nodes scored
per query by $48.9\times$ for flat retrieval and by $6.4\times$ for \sct{}: 386
nodes against 81. Retrieval cost is much closer to independent of collection size,
as the formulation predicts.

We report nodes scored rather than latency for this comparison.
\Cref{tab:main-corpus} lists \sct{} at 7.3\,ms against dense retrieval's 0.4\,ms,
but \sct{} scores residual embeddings in pure Python while the baselines use a
vectorised dot product; that comparison measures implementations. Index
construction cost is in \Cref{tab:cost}.

\paragraph{The metric matters more than the method.}
\Cref{tab:diagnostic} reports token overlap between the retrieved context and the
gold answer, a metric sometimes used as a cheap proxy for retrieval quality. Its
precision term divides by context length, so it rises as context shrinks
regardless of whether the context answers anything: dense retrieval at top-1
scores $3.4\times$ dense retrieval at top-5 on this measure while scoring
\emph{lower} on Answer F1. A system evaluated on context-overlap F1 can therefore
be improved by returning less text. We include the table because the failure mode
is not obvious from the definition, and because any retrieval comparison drawn
from it will be dominated by context length.

\section{Analysis}
\label{sec:analysis}

\paragraph{Why routing fails, and why it is structural.}
Under progressive descent, choosing a document is the first selection the
algorithm makes, and it is made by comparing the query against each document's
root residual. The root residual is by construction the most compressed node in
the tree: one to two sentences standing in for an entire paper. A query asking
which optimiser was used in a specific ablation has almost nothing to match
against in such a summary.

Flat retrieval cannot make this error, because it never summarises before
comparing --- the query is scored against every passage, and a passage mentioning
the optimiser is reachable directly. This is not a tuning deficiency in our
implementation. It is a consequence of ordering compression before selection, and
it applies to any strictly top-down index over a collection.

The controlled comparison is \sct{}-Flat, which fixes the residual content and
changes only the retrieval strategy: routing accuracy rises from 0.202 to 0.347
and Answer F1 from 0.122 to 0.147, a paired difference of $-0.0255$ against the
tree ($p = 0.009$, 95\% CI $[-0.047, -0.004]$). The hierarchy, not the residual
representation, is what costs the accuracy.

It is worth being precise about what this does and does not reject. The tree is
indispensable at \emph{construction} time --- a residual is defined against a
parent, so there is no residual without the hierarchy that produces it, and the
residual is the component that wins. What these experiments reject is the
narrower proposition that descending that hierarchy is a good way to
\emph{retrieve} from it.

\paragraph{Why the depth sweep is flat, and why that is informative.}
All depth settings score within 0.0006 F1 of one another under P2, and routing
accuracy is 0.202 for every one of them. Once the first level has selected the
wrong document, no amount of further descent can recover: depth is conditionally
irrelevant given a routing error, and routing errs on four questions in five.
Under P1, where routing cannot fail, depth still buys only $+0.0026$ F1 for a 25\%
increase in context. The multi-resolution argument requires documents deep enough
to have multiple resolutions, and scientific papers largely do not.

\paragraph{Compression strategy: the free option wins.}
\sct{}-LLM underperforms \sct{}-Extractive (0.233 vs.\ 0.274) while costing
substantially more to build. The evidence metrics locate the failure precisely.
Verbatim evidence recall collapses to 0.009 --- essentially no gold paragraph
survives an LLM-compressed residual intact --- while token-level recall holds at
0.590. The compressor is finding roughly the right region and then paraphrasing
away the text that constitutes the evidence. For extractive compression the same
pair is 0.720 and 0.895: it also loses source text, but far less of it.

This is a general point about summarising indexes rather than a fact about one
compressor. Answer-level metrics alone would have shown a modest deficit and
invited a prompt-tuning explanation; the recall pair shows that the mechanism is
lossy paraphrase, which prompt tuning does not fix.

The ordering reverses under the corpus protocol, where \sct{}-LLM beats
\sct{}-Extractive by $0.0154$ ($p = 0.040$). Routing explains the reversal: an
LLM-written root summary is a better description of a paper than two extracted
sentences, so it routes more accurately (0.358 against 0.202) --- and once
document selection is the binding constraint, that advantage outweighs the
evidence the compressor destroys. The two protocols therefore recommend different
compressors, which is itself an argument for reporting both rather than one.

\paragraph{What the residual formulation is buying.}
Against full summaries at each node, residuals gain 34\% relative F1. The
mechanism is visible in the evidence metrics: full summaries reach only 0.152
verbatim and 0.584 token recall against 0.720 and 0.895 for residuals. Storing
what a node \emph{adds} preserves specifics --- numbers, names, dataset sizes ---
that summarising the node in isolation discards, because those specifics are
exactly what a summary judges unimportant relative to the section's main point.
This is the component of the method we would keep.

\paragraph{Where this leaves the design.}
The two findings compose into a concrete recommendation. Residual storage is worth
keeping: it matches dense retrieval at 70\% of the context budget with no LLM
indexing cost. Strictly top-down routing is not: it halves document selection
accuracy for a cost saving the accuracy loss exceeds on this benchmark. The
design these results point to is a hybrid --- select passages flat, then use the
tree for resolution control within the selected document, keeping the residual
representation and discarding top-down routing. We have not evaluated it and make
no claim for it here.

\paragraph{Absolute numbers.}
Answer F1 near 0.27 under P1 is in the range reported for QASPER, which is a hard
benchmark: questions are written from abstracts by readers who have not seen the
full text, and many require synthesis across sections. Exact match is 0.000 for
every system, which is expected rather than anomalous --- a generated sentence
almost never matches a short reference span byte for byte --- and we therefore do
not report it as a discriminating metric.

\section{Discussion}
\label{sec:discussion}

\paragraph{Compression before selection.}
The clearest lesson from these experiments is about ordering. Any index that
summarises before it selects must make its earliest and most consequential
decision from its least informative representation. Progressive descent takes this
to its limit: the choice of document is made from a one-to-two-sentence root
residual, and a wrong choice cannot be recovered at any later level. Flat
retrieval inverts the order --- select first, over full passages --- and pays for it
in per-query work that grows with the collection.

Framed this way, the trade-off is not specific to \sct{}. It applies to RAPTOR's
summary levels, to table-of-contents indexes, and to any router that operates on
document-level abstractions. Our contribution here is a measurement of the price:
on QASPER, moving document selection onto compressed representations halves
routing accuracy, and the resulting quality loss exceeds the $7.6\times$ reduction
in nodes scored that it buys.

\paragraph{Relation to multi-resolution analysis.}
The residual formulation bears a structural analogy to wavelet decomposition,
where each level stores detail coefficients for a frequency band and the
reconstruction formula plays the role of our accumulation property
(\Cref{prop:accumulation}). The analogy is suggestive rather than formal, and the
disanalogy is instructive: wavelet decomposition is invertible, whereas semantic
subtraction is lossy, and our evidence-recall measurements quantify by how much.
An information-theoretic account of semantic compression --- in which that loss is
bounded rather than merely measured --- remains open.

\paragraph{Beyond documents.}
The construction is source-agnostic: our SQL parser maps a database schema to the
same structure (database $\to$ tables $\to$ columns $\to$ row samples), and unlike
prose this hierarchy is exact rather than inferred. Structured sources are also
the case where our negative result may not transfer. A schema hierarchy has
genuine depth and precise node boundaries, and a table name is a far better
routing key than a paper's one-sentence summary. Whether descent is competitive
there is an empirical question we have not answered, and it is the setting we
would test next.

\paragraph{Future directions.}
Three follow-ups are implied directly by these results. \textbf{Hybrid retrieval}:
select passages flat, then descend within the selected document, keeping the
residual representation and discarding top-down routing. \textbf{Better routing
keys}: score a document by its level-one residuals rather than its root, or index
roots with an approximate nearest-neighbour structure, which would also remove the
linear term from the cost bound. \textbf{Deeper corpora}: evaluate where documents
have real depth --- books, standards, legal codes, database schemas --- since more
than half our papers have no subsections for the depth bound to act on.

\section{Conclusion}
\label{sec:conclusion}

We presented Semantic Compression Trees, a hierarchical index in which each node
stores only its semantic residual, together with a progressive descent retrieval
algorithm and an evaluation on QASPER under two protocols that separate index
quality from document routing.

The representation holds up and the retrieval strategy does not. When the relevant
document is supplied, \sct{} with a zero-LLM extractive compressor matches dense
retrieval on answer quality (0.274 vs.\ 0.277, $p = 0.37$) using 30\% fewer context
tokens and no LLM calls to build the index, and residual storage beats storing full
summaries at each node by 34\% relative ($p < 0.001$). Increasing the collection
fifty-fold multiplies flat retrieval's per-query scoring work by $48.9\times$ and
\sct{}'s by $6.4\times$, confirming that descent decouples query cost from
collection size.

Progressive descent, however, contributes nothing over flat retrieval of the same
residuals when the document is given ($p = 0.27$), and loses substantially when the
document must be found (0.122 vs.\ 0.165, $p < 0.001$). The mechanism is document
routing: descent selects the correct paper 20.2\% of the time against 39.3\% for
flat retrieval, because that selection is made from the most compressed node in the
tree. Depth beyond one level is worth $+0.003$ F1 for 25\% more context. The
choice of compressor depends on which protocol applies: the free extractive
variant is the better one when the document is given, since LLM compression
collapses verbatim evidence recall to 0.009, but LLM compression wins under the
corpus protocol because its root summaries route more accurately.

The result we would build on is the residual representation, not the descent.

\section*{Limitations}

\paragraph{Scale and single dataset.}
We evaluate on 50 QASPER papers and 173 questions. That is enough to place
confidence intervals on the differences we report, and the intervals are wide
enough that several comparisons are not resolved at this sample size; we say so
where that is the case rather than reading a point estimate as a result. One
dataset also cannot separate properties of \sct{} from properties of NLP papers,
whose section conventions are unusually regular. NarrativeQA~\citep{kocisky2018narrativeqa},
HotpotQA~\citep{yang2018hotpotqa}, and Natural Questions~\citep{kwiatkowski2019nq}
would each stress a different assumption, and multi-hop questions in particular
would test whether a single descent path is the right shape for evidence that must
be assembled from several branches.

\paragraph{Structure dependence.}
The tree is derived from the document's own headings, so \sct{} inherits whatever
structure the source provides. QASPER papers are close to a best case: explicit,
consistently nested sections. Even here, 28 of 50 papers have no subsections at
all, so the depth bound has nothing to act on for more than half the corpus --- a
ceiling on how much a multi-resolution argument can be demonstrated on this
dataset. Unstructured sources (scanned PDFs, transcripts, chat logs) would fall
back on inferred boundaries, which we do not evaluate.

\paragraph{Residual quality and its cost.}
Semantic subtraction ($\ominus$) is approximated, not exact. Extractive
compression retains sentences that partially restate the parent; LLM compression
can discard novel but low-salience detail. The consequence is measurable rather
than hypothetical: the gap between verbatim and token-level evidence recall
quantifies how much source text a residual fails to preserve, and it is the
clearest cost of the representation. Applications that need exact source wording
should read that gap, not the faithfulness score. A formal measure of residual
quality remains open.

\paragraph{Top-down routing is bounded by the root summary.}
Under the corpus protocol, document selection is the first level of descent and
therefore depends on the root residual, which is by construction the most
aggressively compressed node in the tree --- one to two sentences standing in for
an entire paper. A flat retriever compares the query against every passage and so
cannot make this class of error at all. Whether a top-down index can route as well
as an exhaustive scan is a question about the first level, not about the method's
descent, and a hybrid that routes flat and descends hierarchically is the obvious
thing to try next. We do not attempt it here.

\paragraph{Judge-based metrics.}
The judge is a single model scoring with a rubric, and it is the same model family
used to generate the answers, which risks self-preference. We report judge scores
beside token-level and evidence-level metrics for exactly this reason, and note
where they disagree; we do not treat agreement between them as validation, nor
disagreement as noise.

\paragraph{Latency measurement.}
Reported retrieval latency excludes the query-embedding round trip, which is
served from a cache uniformly for every embedding-based system. This isolates
retrieval computation --- the quantity the cost analysis concerns --- from a network
constant that is identical across those systems, but it means the figures are not
end-to-end serving latency.

\section*{Reproducibility}

\ifcodereleased
\else
Each experiment writes a report recording the code revision that produced it, the
deployment names, the tokeniser, the bootstrap seed, per-question records for every
system, and the measured API usage of the run. Every table in this paper is
generated from those reports by script rather than transcribed, so a table cannot
diverge from the run behind it.

Answer generation and judge scoring depend on a hosted model and are therefore not
bitwise reproducible from scratch. Every API response is cached under a content
hash of its request, and we re-ran both protocols against that cache: the
confirmation pass reissued 58 of roughly 14{,}900 calls and reproduced every
reported figure exactly. Retrieval, tree construction, extractive compression, and
all statistics are deterministic given a fixed seed.
\fi

Two facts about the harness are worth stating explicitly, because they are the
kind of detail that decides whether a retrieval result means anything.
\textbf{First}, systems are driven through an interface that exposes only the
query and, under P1, the document identifier; a reference answer or evidence
annotation cannot reach a retrieval code path, and \ifcodereleased
\else the harness
carries regression tests that assert this, including one that reconstructs an
answer-maximising selector and confirms the tests reject it\fi. \textbf{Second},
retrieval is executed sequentially while generation and judging are parallelised,
because retrieval latency is a reported quantity and would otherwise measure
thread contention.

\section*{Ethics Statement}

\paragraph{Environmental Impact.}
\sct{}-LLM construction issues LLM calls proportional to the number of tree nodes;
the per-run figure is recorded in \Cref{tab:cost}. The extractive variant was
designed and evaluated as a zero-LLM-cost alternative precisely so that indexing
cost is a choice rather than a fixed price, and we report what that choice costs in
retrieval quality rather than assuming it is free.

\paragraph{Faithfulness and Misinformation.}
Any summarising index can distort what it compresses, and \sct{} compresses by
construction: a residual is lossy with respect to its source. We measure this from
two directions---judge-scored faithfulness of generated answers, and the gap
between verbatim and token-level evidence recall, which quantifies how much source
text a residual fails to preserve. For applications where exact source wording
carries legal or clinical weight, that gap, not the faithfulness score, is the
number to read.

\paragraph{Content filtering.}
A small number of questions could not be scored because the answer-generation
provider refused the prompt on content-policy grounds; QASPER indexes NLP papers,
including research on hate speech and offensive language. We report which questions
were excluded and apply the exclusion identically to every system, since the filter
responds to retrieved content and would otherwise penalise the systems that
retrieved the relevant passage.

\paragraph{Data.}
All experiments use the publicly available QASPER dataset~\citep{dasigi2021qasper},
distributed under CC BY 4.0. No private or personally identifiable data was used.

\section*{Acknowledgments}
We thank the QASPER~\citep{dasigi2021qasper} authors for making their dataset publicly available and the Allen Institute for AI for hosting it.

\section*{About the Author}

\textbf{Junaid Farooq} works at the boundary between AI research and production
systems, with over a decade in artificial intelligence, machine learning, and
software engineering.

He holds a PhD in Artificial Intelligence (2023) from the Artificial Intelligence
Laboratory, Department of Electrical Engineering, National Institute of Technology
Srinagar, where his doctoral work developed hybrid deep-learning architectures for
spatiotemporal forecasting. He has written several peer-reviewed papers and two
books: \emph{The Art of Code: Tactics and Principles of Clean Code and
Architecture} (2024), on engineering rigour and maintainable architecture, and
\emph{Zero Day: The Invisible War --- Mastering Vulnerability in the Age of
Artificial Intelligence} (2025), on security and safety for AI systems.

He is Chief Architect and Vice President of Artificial Intelligence at Sprouts.ai,
where he leads platform architecture and production LLM and agentic systems.

\bibliographystyle{plainnat}

\end{document}